\documentclass{article}

\usepackage{main}

\usepackage[colorlinks=true,citecolor=blue]{hyperref}
\usepackage[english]{babel}
\usepackage{graphicx}
\usepackage{amsmath}
\usepackage{amsfonts}
\usepackage{amssymb}
\usepackage{mathtools}
\usepackage{natbib}
\usepackage{mathtools}
\usepackage{algorithm}
\usepackage[noend]{algpseudocode}
\usepackage{amsthm}

\newtheorem{mytheorem}{Theorem}[section]
\newtheorem{cor}{Corollary}[section]
\newtheorem{myproof}{Proof}[section]
\newtheorem{example}{Example}[section]
\newtheorem{definition}{Definition}[section]

\theoremstyle{definition}

\title{Stability of decision trees and logistic regression}

\author{
  Nino Arsov \\
  Macedonian Academy of Sciences and Arts\\
  1000 Skopje, Macedonia \\
  \texttt{narsov@manu.edu.mk} \\
   \And
 Martin Pavlovski \\
  Temple University\\
  Philadelphia, PA 19122 \\
  \texttt{martin.pavlovski@temple.edu} \\
  \And
  Ljupco Kocarev \\
  Macedonian Academy of Sciences and Arts\\
  1000 Skopje, Macedonia\\
  \texttt{lkocarev@manu.edu.mk}
}

\begin{document}

\maketitle

\begin{abstract}
Decision trees and logistic regression are one of the most popular and well-known machine learning algorithms, frequently used to solve a variety of real-world problems. Stability of learning algorithms is a powerful tool to analyze their performance and sensitivity and subsequently allow researchers to draw reliable conclusions. The stability of these two algorithms has remained obscure. To that end, in this paper, we derive two stability notions for decision trees and logistic regression: hypothesis and pointwise hypothesis stability. Additionally, we derive these notions for $L_2$-regularized logistic regression and confirm existing findings that it is uniformly stable. We show that the stability of decision trees depends on the number of leaves in the tree, i.e., its depth, while for logistic regression, it depends on the smallest eigenvalue of the Hessian matrix of the cross-entropy loss. We show that logistic regression is not a stable learning algorithm. We construct the upper bounds on the generalization error of all three algorithms. Moreover, we present a novel stability measuring framework that allows one to measure the aforementioned notions of stability. The measures are equivalent to estimates of expected loss differences at an input example and then leverage bootstrap sampling to yield statistically reliable estimates. Finally, we apply this framework to the three algorithms analyzed in this paper to confirm our theoretical findings and, in addition, we discuss the possibilities of developing new training techniques to optimize the stability of logistic regression, and hence decrease its generalization error.

\end{abstract}

\section{Introduction}
\label{sec:introduction}

As researchers are flocking to machine learning in order to meet the soaring demands from AI-powered businesses, reproducibility of important new results that reverberate across the scientific community is critical for drawing reliable conclusions. Sensitivity analysis of learning algorithms is a powerful tool to assess their performance in real-world applications. In computational learning theory, we generally understand the sensitivity of learning algorithms in terms of their ability to produce consistent outcomes and reliably generalize to unobserved data. One way to quantify the sensitivity of learning algorithms is to investigate their sensitivity to changes in the training set, known as stability.
In a more general sense, researchers can draw reliable conclusions about the performance of ``stable'' learning algorithms when the reproducibility of results comes into question.

The earliest notions of stability date back to the seventies, when Devroye and Wagner investigated the stability of the leave-one-out error with respect to changes in the training set~\citep{devroye1979distribution, devroye1979distribution2}, but the most significant advance in defining different notions of stability of learning algorithms was made by~\cite{elisseeff1}. They introduced different notions of stability that they leveraged for deriving upper bounds on the generalization error of learning algorithms. They defined three important types of stability: hypothesis stability, pointwise hypothesis stability, and uniform stability. The first is the weakest, while the last is the strongest notion of stability. According to~\cite{elisseeff1}, a learning algorithm is stable when its stability scales reciprocally to the size of the training set, i.e., it scales as $O(1/m)$.

As the previous paragraph states, hypothesis stability is the weakest of the three notions. Hypothesis stability has been derived for the $k$-nearest neighbors algorithm ($k$-NN)~\citep{elisseeff1}, for linear regression and regularized logistic regression in a technical report, for the AdaBoost algorithm~\citep{kutin2001interaction}, and for the Gentle Boost algorithm~\citep{arsov2017generating} as a constituent of a collaborative ensemble scheme~\citep{arsov2017generating,pavlovski2018generalization}. L2 regularization with a penalty $\lambda\in\mathbb{R}^+$ stabilizes learning algorithms because their objective functions then become $\lambda$-strongly convex, which leads to meeting the stability criterion.

Another aspect of stability theory is the importance of quantifying and measuring the stability of outcomes produced by learning algorithms. Despite ample research on stability, formal ways of measuring hypothesis and pointwise hypothesis stability have not been established yet. Stability measures allow researchers to estimate and interpret the stability of learning algorithms from different perspectives. \cite{poggio2004general} introduced and measured a different notion of stability, called cross-validation leave-one-out stability; \cite{Kuncheva2007} defined the stability index; later,~\cite{gavinSarah1,gavinSarah2} introduced a variety of stability measures for feature selection algorithms. A recent paper introduced a framework for measuring the stability of the outcomes of any learning algorithm~\citep{philippMeasuring}. This framework comprises two phases: first, two new training sets are resampled from the training sets by either bootstrap sampling, subsampling, or splitting. Then, a separate evaluation set is used to measure the stability of the algorithm by a variety of similarity (and distance) measures. This set either includes all training examples, or the out-of-bag examples as well as the so-called out-of-sample examples that do not appear in any of the two training sets.

To that end, in this paper, we address the lack of hypothesis and pointwise hypothesis stability expressions and means to measure them. We focus on logistic regression~\citep{kmurphy} and decision trees~\citep{quinlan1986induction}, and we extend the work on stability theory and stability measures with the following contributions:
\begin{itemize}
    \item [(1)] we derive hypothesis and pointwise hypothesis stability for logistic regression, including $L_2$ regularization, and decision trees; for logistic regression, we show that its stability depends on the smallest eigenvalue of the Hessian matrix of the cross-entropy loss, while for decision trees, it depends on the number of leaves; we then use these results to derive upper bounds on their generalization error,\\
    \item[(2)] we address the lack of stability measures for hypothesis and pointwise hypothesis stability by adapting the stability framework by~\cite{philippMeasuring} to hypothesis and pointwise hypothesis stability.
\end{itemize}
Then, we support our theoretical findings with experiments using our framework to measure the stability on a synthetic dataset. Although expressions for uniform stability can be easily derived, its strong requirement that the change in the error is very small implies that most algorithms are not uniformly stable, except L2-regularized algorithms, including support vector machines~\citep{poggio2004general}.

Here, for logistic regression, we show that the hypothesis and pointwise hypothesis stability is reciprocal to smallest possible eigenvalue of the Hessian matrix of the cross-entropy loss, and for decision trees, it depends on the number of leaves. In addition, we develop a framework to measure hypothesis and pointwise hypothesis stability. We then use it to confirm our theoretical results.

This paper is organized as follows: in Section~\ref{sec:prerequisites} we give the prerequisites and notation for hypothesis and pointwise hypothesis stability, in Section~\ref{sec:base_stability} we derive the hypothesis and pointwise hypothesis stability of logistic regression and decision trees, in Section~\ref{sec:framework} we present a framework for measuring hypothesis and pointwise hypothesis stability, and in Section~\ref{sec:experiments} we use this framework to support our theoretical findings. We conclude the paper and discuss future work with Section~\ref{sec:conclusion}.

\section{Prerequisites and notation}
\label{sec:prerequisites}

This section briefly introduces the notation used in the rest of the paper and the prerequisite notions that help analyze learning algorithms from the perspective of stability.
First, this paper follows the notation:
\begin{itemize}
\item[$\bullet$] $\mathbb{P}[\cdot]$: probability
\item[$\bullet$] $\mathbf{E}_X[\cdot]$: expected value with respect to $X$
\end{itemize}

Let $\mathcal{X}$ and $\mathcal{Y}$ denote the input and output space of a learning algorithm. Here, we assume that $\mathcal{X} \subseteq V^d$, where $d \geq 1$ and $V$ is a vector space.  In classification, the output space $\mathcal{Y} = \mathcal{C}$, where $\mathcal{C}$ is the set of the possible class labels.
\\
Let $\mathcal{Z}=\mathcal{X} \times \mathcal{Y}$. 
A training set of $m$ examples, drawn i.i.d. from $\mathcal{Z}$, is denoted by $\mathcal{D}=(z_i=(\vec{x}_i, y_i))_{i=1}^m$.  The set of all possible training sets of size $m$ is $\mathcal{Z}^m$.

\noindent
By removing the $i$-th example from $\mathcal{D}$,  we get a new training set of $m-1$ examples,

\begin{equation*}
\mathcal{D}^{\backslash i} = (z_1,\ldots,z_{i-1},z_{i+1},\ldots,z_m).
\end{equation*}

\noindent
Replacing the $i$-th example from $\mathcal{D}$ with some $z \in\mathcal{Z}$ and $z \notin \mathcal{D}$ yields

\begin{equation*}
\mathcal{D}^i = \{z_1,\ldots,z_{i-1},z,z_{i+1},\ldots,z_m\},\quad z \in \mathcal{Z},z \notin \mathcal{D}.
\end{equation*}

\noindent
We denote the outcome of a learning algorithm trained on $\mathcal{D}$ by $f_{\mathcal{D}}$. The loss function of $f_{\mathcal{D}}$ with respect to $z$ is $\ell(f_{\mathcal{D}},z)$.

Throughout this paper, we use the following loss functions:

\begin{enumerate}
\item Classification loss $\ell(f,z)\in\{0,1\},$ $$\ell(f,z)=I(f(\vec{x})\neq y),$$ where $I(C)$ is an indicator function equal to 1 when $C$ is true, and equal to 0 otherwise.
\\
\item $\gamma$-loss $\ell_{\gamma}\in\left[0,1\right],$
$$\ell_{\gamma}(f, z) =\begin{cases}
1, & \textrm{ if } yf(\vec{x}) < 0,\\
1 - \displaystyle \frac{yf(\vec{x})}{\gamma}, & \textrm{ if }  0\leq yf(\vec{x})\leq \gamma,\\
0, & \textrm{ otherwise}.
\end{cases}$$ 
\end{enumerate}
In  addition we take into account the loss function minimized by a learning algorithm. Such example is the cross-entropy loss used in logistic regression.

\begin{definition}[Hypothesis stability]
A learning algorithm has hypothesis stability $\beta_h(m)$ with respect to a loss function $\ell$ if the following holds:
\begin{equation*}
\forall i \in \{1,2,\ldots,m\},\quad \mathbf{E}_{\mathcal{D},z}\left[|\ell(f_{\mathcal{D}},z) - \ell(f_{\mathcal{D}^{\backslash i}},z)|\right] \leq \beta_{h}(m).
\label{eq:hypothesis_stability}
\end{equation*}
\label{def:hypothesis_stability}
\end{definition}

\begin{definition}[Pointwise hypothesis stability]
A learning algorithm has pointwise hypothesis stability $\beta_{ph}(m)$ with respect to the loss function $\ell$ if the following holds:

\begin{equation*}
\forall i \in \{1,2,\ldots,m\},\quad \mathbf{E}_{\mathcal{D}}\left[|\ell(f_{\mathcal{D}},z_i) - \ell(f_{\mathcal{D}^{\backslash i}\cup z},z_i)|\right] \leq \beta_{ph}(m).
\label{eq:pointwise_hypothesis_stability}
\end{equation*}
\label{def:pointwise_hypothesis_stability}
\end{definition}

\begin{definition}[Uniform stability]
An algorithm has uniform stability $\beta_u(m)$ with respect to the loss function $\ell$ if the following holds,
\begin{equation*}
\forall \mathcal{D} \in \mathcal{Z}^m,\, \forall i\in\{1,\ldots,m\},\quad \| \ell(f_{\mathcal{D}},\, .) - \ell(f_{\mathcal{D}^{\backslash i}},\, .)\|_{\infty} \leq \beta.
\label{eq:uniform_stability}
\end{equation*}
\label{def:uniform_stability}
\end{definition}

\begin{definition}[Generalization error]
The true (generalization) error of a learning algorithm whose outcome is $f_{\mathcal{D}}$ is 
\begin{equation*}
    R_{gen}(f_{\mathcal{D}}) = \mathbf{E}_z[\ell(f_{\mathcal{D}},z)].
    \label{eq:gen_err}
\end{equation*}
\label{def:gen_err}
\end{definition}

\noindent
The simplest estimator of the generalization error is the empirical error on $\mathcal{D}$.

\begin{definition}[Empirical error]
The empirical error of a learning algorithm whose outcome is $f_{\mathcal{D}}$ is 
\begin{equation*}
    R_{emp}(f_{\mathcal{D}}) = \frac{1}{m}\sum_{i=1}^m\ell(f_{\mathcal{D}},z).
    \label{eq:emp_err}
\end{equation*}
\label{def:emp_err}
\end{definition}

\section{Hypothesis and pointwise hypothesis stability of logistic regression and decision trees}
\label{sec:base_stability}

In this section, we present the hypothesis stability of decision trees and logistic regression, which are often used as the base algorithms in ensemble and stacking~\citep{wolpert1992stacked} settings. We first give an example for the $k$-NN algorithm, borrowed from~\cite[Example 1]{elisseeff_randomized}.

\begin{example}{\textup{\textbf{(Hypothesis stability of $k$-NN)}~\citep[Example 1]{elisseeff_randomized}.}}

\noindent
Let $f_{\mathcal{D}}$ be the outcome of a $k$-NN algorithm on a training set $\mathcal{D}$ with $m$ elements. With respect to the classification loss $\ell$, $k$-NN is at least $k/m$-stable,

\begin{equation}
\mathbf{E}_{\mathcal{D},z}\left[|\ell(f_{\mathcal{D}}, z) - \ell(f_{\mathcal{D}^{\backslash i}}, z)| \right] \leq \frac{k}{m}.
\label{eq:knn_hypothesis_stability}
\end{equation}
\begin{myproof}
\noindent
Let $v_i$ be the neighborhood of $z_i$ such that the
closest example in the training set to any example of $v_i$ is $z_i$, $$v_i=\{z\,|\,\forall z',\text{dist}(z_i, z) < \text{dist}(z', z) \}.$$
In other words, $z_i$ is the 'nearest neighbor' of every $z\in v_i$. 

\noindent
The outcome of the 1-NN algorithm is given by $$f_{\mathcal{D}}(\vec{x}) = \sum_{i=1}^m y_i\mathbf{1}_{\vec{x}\in v_i}(\vec{x}).$$ The difference $|\ell(f_{\mathcal{D}},z) - \ell(f_{\mathcal{D}^{\backslash i}}, z)|$ is then defined by the set $v_i$, given that $\ell$ is the classification loss. We thus have 

\begin{equation*}
\mathbf{E}_{z}[|\ell(f_{\mathcal{D}},z) - \ell(f_{\mathcal{D}^{\backslash i}}, z)|] \leq \mathbb{P}(v_i).
\end{equation*}

\noindent
Also, $v_i$ depends on $\mathcal{D}$. To compute hypothesis stability, we need $\mathbf{E}_{\mathcal{D},z}$, and by taking $\mathbf{E}_{\mathcal{D}}$ on both sides in the equation above to compute the hypothesis stability,

\begin{equation}
\mathbf{E}_{\mathcal{D},z}[|\ell(f_{\mathcal{D}},z) - \ell(f_{\mathcal{D}^{\backslash i}}, z)|] \leq \mathbf{E}_{\mathcal{D}}[\mathbb{P}(v_i)].
\label{eq:hyp_stb_knn}
\end{equation}

\noindent
Averaging over $\mathcal{D}$, we need to compute $\mathbf{E}_{\mathcal{D}} [\mathbb{P}(v_i)]$, which is the same for all $i$ because every $z_i$ is drawn i.i.d. from the same distribution. But, since $f_{\mathcal{D}}(\vec{x})\in\{-1, 1\}$, we have,
$$
1=\mathbf{E}_{\mathcal{D},z}[|f_{\mathcal{D}}(\vec{x})|]=
\mathbf{E}_{\mathcal{D},z}\left[\left|\sum_{i=1}^m y_i\mathbf{1}_{\vec{x} \in v_i}(\vec{x})\right|\right] = \mathbf{E}_{\mathcal{D},z}\left[\mathbf{1}_{\vec{x}\in v_j}(\vec{x})\right],\quad 1\leq j \leq m.
$$

\noindent
The last equality comes from the fact that for fixed $\mathcal{D}$ and $z$, only one $\mathbf{1}_{\vec{x}\in v_i} (\vec{x})$ is non-zero. We also
have that

\begin{equation}
1=\mathbf{E}_{\mathcal{D},z}\left[\mathbf{1}_{x\in v_i}(\vec{x})\right]=m\mathbf{E}_{\mathcal{D}}[\mathbb{P}(v_i)].
\label{eq:hyp_stb_knn_1}
\end{equation}

\noindent
Consequently, $\mathbf{E}_{\mathcal{D}}[\mathbb{P}(v_i)]=1/m$ and, putting Equations~\eqref{eq:hyp_stb_knn}~and~\eqref{eq:hyp_stb_knn_1} together, $1$-NN has hypothesis stability bounded above by $1/m$, that is,
$$
\mathbf{E}_{\mathcal{D},z}[|\ell(f_{\mathcal{D}},z) - \ell(f_{\mathcal{D}^{\backslash i}}, z)|] \leq \frac{1}{m}.
$$

\noindent
Finally, from Equation~\eqref{eq:hyp_stb_knn_1}, $k$-NN has hypothesis stability bounded above by $k/m$.
\qed
\end{myproof}
\label{ex:1}
\end{example}

\subsection{Hypothesis and pointwise hypothesis stability of decision trees}

In this part, we derive the hypothesis stability and pointwise hypothesis stability of decision trees. We prove that both kinds of stability depend on the number of leaves, or, in other words, the depth of the tree.

\begin{mytheorem}[Hypothesis stability of decision trees]
\label{thm:dt_stability}
With respect to the classification loss, a decision tree with $v$ leaves, induced from a training set $\mathcal{D}$, has hypothesis stability bounded above by $1/v$,

\begin{equation*}
\mathbf{E}_{\mathcal{D},z}\left[|\ell(f_{\mathcal{D}}, z) - \ell(f_{\mathcal{D}^{\backslash i}}, z)|\right] \leq \frac{1}{v}.
\label{eq:dt_stability}
\end{equation*}
\begin{myproof}

$ $\newline
Consider a binary classification setting where $\mathcal{Y}=\{-1, 1\}$. The proof can be extended easily to fit to a multiclass setting. Here, we focus on the former.

\noindent
The decision tree algorithm partitions the training set $\mathcal{D}$ into $v \leq m$ disjoint subsets, each being a leaf at the bottom of the tree. Each $z_i \in \mathcal{D}$ appears in exactly one leaf.
Let $s$ be the branching factor of the tree, that is, the largest possible number of children per node. 
The splitting criterion at each node $u$ in the tree is a predicate over the features of $\mathcal{D}$, a tuple $c_u=(attribute, operator, value)$.
Let $h$ be the depth (height) of the tree, such that $h\leq \log_s m$. As the tree becomes deeper, the number of leaves $v$ gets closer to $m$, and if $v = m$, then each example in $\mathcal{D}$ forms a leaf node and $h=\log_s m$.

\noindent
Let $\mathcal{L}_{\mathcal{D}}=\{l_k\,|\,l_k \subset \mathcal{D}\}_{k=1}^v$ be the set of leaf nodes. Each leaf node $l_k$ contains at least one example from $\mathcal{D}$, one being the case when $v = m$. Every example placed in $l_k$
The subtree rooted at a node $u\notin \mathcal{L}_{\mathcal{D}}$ contains a subset of examples $\mathcal{D}(u)\subseteq\mathcal{D}$, distributed among its leaf nodes. 
An example $z \in \mathcal{Z}$ is classified by testing whether the feature vector $\vec{x}$ satisfies the sequence of predicates that appear along the path from the root of the tree to one of the leaf nodes $l_k$, for $k\in\{1,2,\ldots,v\}$, which we denote by $x \in \mathcal{D}(l_k)$, or equivalently, $x \vDash c_{l_k}$ and $x \vDash u'$, where $u'$ is a predecessor of $l_k$. The outcome $f_{\mathcal{D}}$ of the algorithm equals the class of the majority of examples in $\mathcal{D}(l_k)$, 

\begin{equation*}
f_{\mathcal{D}}(\vec{x}) = \sum_{j=1}^v \mathbf{1}_{\vec{x}\, \vDash\, l_j} \text{sign}\left(\sum_{z'\in\mathcal{D}(l_j)} y'\right).
\label{eq:decision_tree_outcome}
\end{equation*}

\noindent
For notation brevity, let $l_{z_i}$ denote the leaf node that contains $z_i\in\mathcal{D}$, i.e., $z_i \in \mathcal{D}(l)$. The difference of the losses, $|\ell(f_{\mathcal{D}}, z) - \ell(f_{\mathcal{D}^{\backslash i}}, z)|$ is defined by $\mathbb{P}(l_{z_i})$ because $z_i$ appears in only one of the leaf nodes,and that is $l_{z_i}$. The probability $\mathbb{P}(l_{z_i})$ depends on $\mathcal{D}$, as well. We thus have 

\begin{equation}
\mathbf{E}_z\left[|\ell(f_{\mathcal{D}},z) - \ell(f_{\mathcal{D}^{\backslash i}},z) |\right] \leq \mathbb{P}(l_{z_i}).
\label{eq:hyp_stb_p}
\end{equation}

\noindent
The next step is taking the average over $\mathcal{D}$ to make the left-hand side of Equation~\eqref{eq:hyp_stb_p} satisfy the hypothesis stability definition, 

\begin{equation*}
\mathbf{E}_{\mathcal{D},z}\left[|\ell(f_{\mathcal{D}},z) - \ell(f_{\mathcal{D}^{\backslash i}},z) |\right] \leq \mathbf{E}_{\mathcal{D}}[\mathbb{P}(l_{z_i})].
\label{eq:dt_hyp_stb}
\end{equation*}

\noindent
The goal at this point is to compute $\mathbf{E}_{\mathcal{D}}[\mathbb{P}(l_{z_i})]$, for which we use a technique similar to the one in Ex.~\ref{ex:1}; since every $z_i$ is drawn i.i.d. from an unknown distribution, and $\mathcal{D}$ is partitioned into $v$ disjoint subsets at the bottom of the tree, the expected value $\mathbf{E}_{\mathcal{D}}[\mathbb{P}(l_{z_i})]$ is the same for all $v$ leaf nodes under the assumption that all data features are independent and identically distributed. Given that for any $\vec{x}$,  $f_{\mathcal{D}}(\vec{x}) \in \{-1, 1\}$, we have, on one hand,

\begin{equation*}
1=\mathbf{E}_{\mathcal{D},z}\left[|f_{\mathcal{D}}(\vec{x})|\right] = \mathbf{E}_{\mathcal{D},z}\left[\left|\sum_{j=1}^v \mathbf{1}_{\vec{x}\, \vDash \, l_j} \text{sign}\left(\sum_{z_i\in\mathcal{D}(l_j)}y_i\right)\right|\right] = \mathbf{E}_{\mathcal{D},z}\left[\mathbf{1}_{\vec{x}\, \vDash\, l_k}\right],
\end{equation*}

\noindent
the reason being that only one $\mathbf{1}_{\vec{x}\,\vDash\,l_j}$ is non-zero for $j=k$, while the $\text{sign}(\,.\,)$ is irrelevant. On the other hand, $$1=\mathbf{E}_{\mathcal{D},z} \Big[\sum_{j=1}^v \mathbf{1}_{\vec{x}\,\vDash \, l_j}\Big]=v\mathbf{E}_{\mathcal{D}}[\mathbb{P}(l_{z_i})]$$ because $\sum_{j=1}^v\mathbf{1}_{\vec{x}\,\vDash\,l_j}$ is the whole probability space of $\mathbb{P}(l_{z_j})$, thus  $\mathbf{E}_{\mathcal{D}}[\mathbb{P}(l_{z_i})]=1/v$. Finally, the hypothesis stability of decision trees for classification scales as $O(1/v)$, or $$\mathbf{E}_{\mathcal{D},z}\left[|\ell(f_{\mathcal{D}},z) - \ell(f_{\mathcal{D}^{\backslash i}}, z)| \right]\leq \frac{1}{v}.$$ For multiclass classification problems, the proof can be reproduced for each class separately.
\qed
\end{myproof}

\end{mytheorem}

In terms of stability, decision trees are similar to the $k$-NN algorithm. When a leaf node is reached after traversing the tree, the training examples in the leaf are equivalent to the $k$ nearest neighbors. On average, a leaf node contains $m/v$ examples. Taking $k=m/v$ and plugging it into Equation~\eqref{eq:knn_hypothesis_stability} results in $m/mv=1/v$, which asserts the proof.
Moreover, $\mathbb{P}(l_{z_i})$ decreases as the tree gets deeper, which follows from the fact that $\mathbb{P}(l_{z_i})$ is the joint probability that $z_i$ has to satisfy every node predicate that appears along the path from the root node $u_0$ to the leaf node $l_{z_i}$. Let this path be $(u_0,\ldots, l_{z_i})$ and let $(c{u_0}, \ldots, c_{l_{z_i}})$ be their associated predicates. Consider the following inequality:

\begin{flalign*}
\mathbb{P}(l_{z_i}) &\approx \mathbb{P}(\vec{x}_i \vDash c_{u_0}, \ldots , \vec{x}_i \vDash c_{l_{z_i}}) \nonumber \\
&= \mathbb{P}(\vec{x}_i \vDash c_{u_0})\, \mathbb{P}(\vec{x}_i \vDash c{u_1}\,|\, \vec{x}_i \vDash c{u_0})\ldots \mathbb{P}(\vec{x}_i \vDash c_{l_{z_i}}\,|\, \vec{x}_i \vDash c_{u_0}, \ldots, \vec{x}_i \vDash c_{u_{h-1}}) \nonumber \\
&\leq \mathbb{P}(\vec{x}_i \vDash c_{u_0})\, \mathbb{P}(\vec{x}_i \vDash c{u_1}\,|\, \vec{x}_i \vDash c{u_0})\ldots \mathbb{P}(\vec{x}_i \vDash c_{u_{h-1}}\,|\, \vec{x}_i \vDash c_{u_0}, \ldots, \vec{x}_i \vDash c_{u_{h-2}})\nonumber \\
&\approx \mathbb{P}(u_{h-1}).
\end{flalign*}

\noindent
When the leaf node containing $z_i$ is removed from the tree, $z_i$ is transferred to $l_{z_i}$'s parent, $u_{h-1}$, i.e., $z_i \in \mathcal{D}(u_{h-1})$ and the number of leaf nodes becomes $v-1$. Consequently, the joint probability $\mathbb{P}(l_{z_i})$ increases to $\mathbb{P}(u_{h-1})$, making the tree less stable at $1/(v-1)$. Put in another way, balanced and deeper decision trees are more stable.

We are interested in $\mathbb{P}(l_{z_i})$, rather than $1-\mathbb{P}(l_{z_i})$, because if $z_i$ is more likely to appear in $\mathcal{D}(l_{z_i})$, then removing $z_i$ from $\mathcal{D}$ is more likely to cause stronger perturbations, and consequently make $f_\mathcal{D}$ more sensitive to changes in $\mathcal{D}$ and with respect to the difference of $\ell(f_{\mathcal{D}},z)$ and $\ell(f_{\mathcal{D}^{\backslash i}}$.
Moreover, $\mathbb{P}(l_{z_i})$ is inversely proportional to the depth $h$, which comes from the fact that $\mathbb{P}(l_{z_i})$ can be expressed by means of the joint probability of all splitting criteria on the path from the root to $l_{z_i}$. Let this path be $(u_0,u_1,\ldots,u_{h-1},l_{z_i})$ and $(p_{u_0},p_{u_1},\ldots,p_{u_{h-1}},p_{l_{z_i}})$ their associated feature predicates. It follows that 
\begin{flalign*}
\mathbb{P}(l_{z_i})&\sim p(c_{u_0},c_{u_1},\ldots,c_{l_{z_i}}) = p(c_{u_0})p(c_{u_1}\,|\,c_{u_0})\cdot\ldots\cdot p(c_{l_{z_i}}\,|\,c_{u_0},c_{u_1},\ldots,c_{u_{h-1}})\nonumber\\
&\leq p(c_{u_0})p(c_{u_1}\,|\,c_{u_0})\cdot\ldots\cdot p(c_{u_{h-1}}\,|\,c_{u_0},c_{u_1},\ldots,c_{u_{h-2}}) = p(c_{u_0},c_{u_1},\ldots,c_{u_{h-1}}).
\end{flalign*}
A point this equation makes is that if $l_{z_i}$ is removed from the path (its depth is decreased by one), and $u_{h-1}$ is now labeled $l_{z_i}$, the joint probability $\mathbb{P}(l_{z_i})$ is consequently increased, making the tree less stable. Also note that in this case the number of leaves $l$ is also decreased.
Put in another way, trees with a larger depth have a better hypothesis stability. Calculating the joint probability $\mathbb{P}(p_{u_0},p_{u_1},\ldots,p_{l_{z_i}})$ is rather challenging since the distribution of $\mathcal{Z}$ is unknown. Therefore, we use a technique similar to the one used in Example~\ref{ex:1}.

Note that, because the $z_i$ are drawn i.i.d. from the distribution, and in contrast to $k$-NN (see Example~\ref{ex:1}), $\mathcal{D}$ is now partitioned into $l$ disjoint subsets, the expectation
$\mathbf{E}_{\mathcal{D}}[\mathbb{P}(l_{z_i})]$ is the same for all $z_i$, i.e., for all $v$ leaves (partitions of $\mathcal{D}$), assuming the data features are independent and identically distributed.
The goal at this point is to compute $\mathbf{E}_{\mathcal{D}}[\mathbb{P}(l_{z_i})]$.

Given that $f_{\mathcal{D}}(\vec{x})\in\{-1,1\}$, we have
\begin{equation*}
1=\mathbf{E}_{\mathcal{D}, z}[|f_{\mathcal{D}}(\vec{x})|] =\mathbf{E}_{\mathcal{D},z}\left[\left|\sum_{k=1}^v \mathbf{1}_{\vec{x}\vDash L_k} \textrm{sign}\Big(\sum_{z_i\in\mathcal{D}(L_k)}y_i\Big)\right|\right] 
=\mathbf{E}_{\mathcal{D},z}[\sum_{k=1}^v \mathbf{1}_{\vec{x}\vDash L_k} ]
\end{equation*}
Here, only one $\mathbf{1}_{\vec{x}\vDash L_k}$ is non-zero since we have a tree, and not a graph. Furthermore, the last equality comes also from the fact that $|\textrm{sign}(\, .\,)|=1$\\
On the other hand, we have that $\sum_{k=1}^l \mathbf{1}_{\vec{x}\vDash L_k} =v\mathbf{E}_{\mathcal{D}}[\mathbb{P}(l_{z_i})]$ because $\sum_{k=1}^v \mathbf{1}_{\vec{x}\vDash L_k}$ defines the whole probability space of $\mathbb{P}(l_{z_i})$, which, being the same for all $l$ leaves results in
\begin{equation*}
1=\mathbf{E}_{\mathcal{D},z}[\sum_{k=1}^v \mathbf{1}_{\vec{x}\vDash L_k} ]=v\mathbf{E}_{\mathcal{D}}[\mathbb{P}(l_{z_i})],
\end{equation*}
and consequently, $\mathbf{E}_{\mathcal{D}}[\mathbb{P}(l_{z_i})]=1/v$. Therefore, a decision tree classifier is 
$1/v$-stable, i.e., its hypothesis stability scales as $O(1/v)$.

In multiclass classification, the proof can be reproduced for each class separately. Existing work on stability in~\citep{elisseeff1,elisseeff_randomized} assumes binary classification as well.
That decision trees are not considered stable unless balanced and considerably large is an immediate consequence of Theorem~\ref{thm:dt_stability}. Since in practice $l$ is often much smaller than $m$ (especially when the tree is pruned), this results in worsened stability. One-level decision or regression trees, called decision stumps, have only a root and two leaves and their stability is only $1/2$. They are a typical example of an unstable algorithm and can be very effectively stabilized using ensemble methods such as bagging or boosting. Theorem~\ref{thm:dt_stability} also reveals that

\begin{itemize}
\item[$\bullet$] through stability, in essence, a decision tree is just a modified $k$-NN algorithm, differing merely in what is considered ``nearest'' -- the points in a leaf or a subset of $\mathcal{D}$

\item[$\bullet$] stability of decision trees deteriorates in a similar fashion to that of the $k$-NN algorithm -- increasing $k$ results in a less stable algorithm since the set of nearest points becomes larger, whereas reducing the number of leaves results in larger disjoint subsets that partition $\mathcal{D}$, hence the deterioration of stability

\item[$\bullet$] larger trees tend to be more stable -- the larger the number of leaves $v$, the better

\item[$\bullet$] similarly, the deeper the tree, the better, because $v\leq s^h$, the root being at level zero; therefore, stability improves exponentially with the depth $h$ 

\item[$\bullet$] balanced trees are considered more stable because then they contain more leaves, i.e., $v=s^h$

\end{itemize}

\subsection{Hypothesis and pointwise hypothesis stability of logistic regression}

In this part, we derive the hypothesis stability and pointwise hypothesis stability of logistic regression. We prove that both kinds of stability are equivalent for this learning algorithm. They depend on the size $m$ of the training set $\mathcal{D}$ as well as the smallest eigenvalue of the Hessian matrix of the cross-entropy loss function at the optimized parameters $\hat{\theta}$.

\begin{mytheorem}[Hypothesis and pointwise hypothesis stability of $L_2$-regularized logistic regression]
Let $f_{\mathcal{D}}$ be the outcome of $L_2$-regularized logistic regression where $\lambda$ is the $L_2$ penalty. Assuming that $\|\vec{x}\|\leq Q$ for all $z\in\mathcal{Z}$ and that the loss function $\ell$ is $\tau$-Lipschitzian, $L_2$-regularized logistic regression has hypothesis (respectively pointwise hypothesis stability
\begin{equation*}
    \forall i\in\{1,2,\ldots m\}\quad \mathbf{E}_{\mathcal{D}, z}\left[|\ell(f_{\mathcal{D}},z) - \ell(f_{\mathcal{D}^{\backslash i}},z)| \right] \leq \frac{2\rho\tau}{\lambda m}Q,
\label{eq:lrl2_hypothesis_stability}
\end{equation*}
\begin{equation*}
    \forall i\in\{1,2,\ldots m\}\quad \mathbf{E}_{\mathcal{D}}\left[|\ell(f_{\mathcal{D}},z_i) - \ell(f_{\mathcal{D}^{\backslash i}},z_i)| \right] \leq \frac{2\rho\tau}{\lambda m}Q.
\label{eq:lrl2_pointwise_hypothesis_stability}
\end{equation*}
\begin{myproof}
The cross-entropy loss function, defined as
\begin{equation*}
\mathcal{L}_{\mathcal{D}}(\theta, \lambda) =- \frac{1}{m}\sum_{i=1}^m y_i\log p(y_i|\vec{x}_i;\theta) + (1-y_i)\log(1 - p(y_i|\vec{x}_i;\theta)) +  \frac{\lambda}{2}\|\theta\|^2,
\end{equation*}
is $\lambda$-strongly convex and therefore, for all $\theta$, $\nabla^2\mathcal{L}(\theta, \lambda)\succeq \lambda$. From the second-order Taylor expansion of $\mathcal{L}_{\mathcal{D}}(\theta, \lambda)$,
$$
\mathcal{L}_{\mathcal{D}}(\hat{\theta}_{\backslash i})\approx \mathcal{L}_{\mathcal{D}}(\hat{\theta}) + (\hat{\theta}_{\backslash i} - \hat{\theta})^T\nabla \mathcal{L}_{\mathcal{D}}(\hat{\theta}) + \frac{1}{2}(\hat{\theta}_{\backslash i} - \hat{\theta})^T\vec{H}(\hat{\theta})(\hat{\theta}_{\backslash i} - \hat{\theta}),
$$
in which the first-order term vanishes because the gradient around the maximum is close to zero, it follows that
\begin{equation}
    \mathcal{L}_{\mathcal{D}}(\hat{\theta}_{\backslash i}, \lambda) - \mathcal{L}_{\mathcal{D}}(\hat{\theta}, \lambda)\geq\frac{\lambda}{2}\|\hat{\theta}_{\backslash i} - \hat{\theta}\|^2,
    \label{eq:l2_ineq}
\end{equation}
since for any $\lambda$-strongly convex function, $(\hat{\theta}_{\backslash i} - \hat{\theta})^T \vec{H}(\hat{\theta})(\hat{\theta}_{\backslash i} - \hat{\theta})\geq \frac{\lambda}{2}\|\hat{\theta}_{\backslash i} - \hat{\theta}\|^2.$
If we take $\Tilde{\mathcal{L}}_{\mathcal{D}}(\theta,\lambda)=-\sum_{i=1}^m y_i \log p(y_i|\vec{x}_i;\theta)+(1-y_i)\log (1-p(y_i|\vec{x}_i;\theta)) + \frac{\lambda m}{2}\|\theta\|^2$, and if $\Tilde{\mathcal{L}}_{\mathcal{D}}(\theta,\lambda)$ is $\rho$-Lipschitzian with respect to its first argument, we can then rewrite Equation~\eqref{eq:l2_ineq} as
\begin{equation}
    \|\hat{\theta}_{\backslash i} - \hat{\theta}\|^2 \leq \frac{2}{\lambda m}\left(\Tilde{\mathcal{L}}_{\mathcal{D}}(\hat{\theta}_{\backslash i},\lambda) - \Tilde{\mathcal{L}}_{\mathcal{D}}(\hat{\theta}, \lambda)\right) \leq \frac{2\rho}{\lambda m}\|\hat{\theta}_{\backslash i} - \hat{\theta}\|.
    \label{eq:l2_ineq_2}
\end{equation}
It is important to note that since $\mathcal{L}_{\mathcal{D}}(\hat{\theta}_{\backslash i}, \lambda) > \mathcal{L}_{\mathcal{D}}(\hat{\theta},\lambda)$, it also holds that $\mathcal{L}_{\mathcal{D}}(\hat{\theta}_{\backslash i}, \lambda) - \mathcal{L}_{\mathcal{D}}(\hat{\theta},\lambda) = |\mathcal{L}_{\mathcal{D}}(\hat{\theta}_{\backslash i}, \lambda) - \mathcal{L}_{\mathcal{D}}(\hat{\theta})|$
Taking $\sigma(x)=1/(1+e^{-x})$ and $\ell$ to be $\tau$-Lipschitzian, we can bound the absolute loss difference (assuming $\|\vec{x}\|\leq Q$):
\begin{flalign}
|\ell(f_{\mathcal{D}}, z) - \ell(f_{\mathcal{D}^{\backslash i}}, z)| &\leq \tau|f_{\mathcal{D}}(\vec{x}) - f_{\mathcal{D}^{\backslash i}}(\vec{x})|\nonumber\\
&= \tau|\sigma(\hat{\theta}^{T}\vec{x}) - \sigma(\hat{\theta}_{\backslash i}^{T}\vec{x})|\nonumber\\
&\leq \tau|\hat{\theta}^T\vec{x} - \hat{\theta}_{\backslash i}\vec{x}|\nonumber\\
&= \tau|(\hat{\theta} - \hat{\theta}_{\backslash i})^T\vec{x}|\nonumber\\
&\leq\tau \|\hat{\theta} - \hat{\theta}_{\backslash i}\|\|\vec{x}\|\nonumber\\
& \leq \tau Q\|\hat{\theta} - \hat{\theta}_{\backslash i}\| \leq \frac{2\rho \tau }{\lambda m}Q.
\label{eq:lrl2_derivation}
\end{flalign}
The last inequality follows from Equation~\eqref{eq:l2_ineq_2}.
Taking $\mathbf{E}_{\mathcal{D},z}[\cdot]$ on both sides of Equation~\eqref{eq:lrl2_derivation} completes the proof for hypothesis stability; replacing $z$ with $z_i\in\mathcal{D}$ completes the proof for pointwise hypothesis stability.
\qed
\end{myproof}
\label{thm:lrl2_hypothesis_stability}
\end{mytheorem}
Moreover, Equation~\eqref{eq:lrl2_derivation} holds for all training sets $\mathcal{D}$, all $i$, and all $z$. Replacing $\mathcal{D}^{\backslash i}$ with $\mathcal{D}^i$ and taking $\mathbf{E}_{\mathcal{D}}$ gives the uniform stability of $L_2$-regularized logistic regression.

\begin{mytheorem}[Hypothesis and pointwise hypothesis stability of logistic regression]
Let $f_{\mathcal{D}}$ be the outcome of logistic regression. Assuming that $\|\vec{x}\|\leq Q$ for all $z\in\mathcal{Z}$ and that the loss function $\ell$ is $\tau$-Lipschitzian, logistic regression has hypothesis (respectively pointwise hypothesis stability
\begin{equation*}
    \forall i\in\{1,2,\ldots m\}\quad \mathbf{E}_{\mathcal{D}, z}\left[|\ell(f_{\mathcal{D}},z) - \ell(f_{\mathcal{D}^{\backslash i}},z)| \right] \leq \frac{2\rho\tau}{m \inf\limits_{\|v\|=1}v^T\vec{H}(\hat{\theta})v}Q,
\label{eq:lr_hypothesis_stability}
\end{equation*}
\begin{equation*}
    \forall i\in\{1,2,\ldots m\}\quad \mathbf{E}_{\mathcal{D}}\left[|\ell(f_{\mathcal{D}},z_i) - \ell(f_{\mathcal{D}^{\backslash i}},z_i)| \right] \leq \frac{2\rho\tau}{m \inf\limits_{\|v\|=1}v^T\vec{H}(\hat{\theta})v}Q,
\label{eq:lr_pointwise_hypothesis_stability}
\end{equation*}
where $\hat{\theta}$ is the minimizer of the cross-entropy loss $\mathcal{L}_{\mathcal{D}}(\theta)$ on $\mathcal{D}$ and $\vec{H}(\hat{\theta})=\nabla^2\mathcal{L}_{\mathcal{D}}(\hat{\theta}).$
\begin{myproof}
This can be proven in the same way as Theorem~\ref{thm:lrl2_hypothesis_stability}. Let $\Tilde{\mathcal{L}}_{\mathcal{D}}(\theta) = -\sum_{i=1}^m y_i\log p(y_i|\vec{x}_i;\theta) + (1-y_i)\log(1-p(y_i|\vec{x}_i;\theta))$.
In this case, $\vec{H}(\theta)\succ0$ for all $\theta$ since $\mathcal{L}_{\mathcal{D}}$  is strictly convex. But, trivially, for $\theta=\hat{\theta}$, $\vec{H}(\theta)\succ \lambda_1(\vec{H}(\theta))$, where $\lambda_1\in\mathbb{R},\lambda_1>0$ is the smallest eigenvalue. 

\begin{equation}
\|\hat{\theta}_{\backslash i} - \hat{\theta}\|^2 \leq \frac{2}{m\lambda_1(\vec{H}(\hat{\theta}))}\left(\Tilde{\mathcal{L}}_{\mathcal{D}}(\hat{\theta}_{\backslash i})-\Tilde{\mathcal{L}}_{\mathcal{D}}(\hat{\theta})\right)\leq \frac{2\rho}{m\lambda_1(\vec{H}(\hat{\theta}))}\|\hat{\theta}_{\backslash i} - \hat{\theta}\|.
\label{eq:lr_ineq}
\end{equation}
Applying this inequality to Equation~\eqref{eq:lrl2_derivation} we get
\begin{equation*}
    |\ell(f_{\mathcal{D}}, z) - \ell(f_{\mathcal{D}^{\backslash i}}, z)| \leq \frac{2\rho\tau}{\lambda_1(\vec{H}(\hat{\theta}))m}
\end{equation*}
For hypothesis stability, we have
\begin{equation}
    \mathbf{E}_{\mathcal{D},z}\left[|\ell(f_{\mathcal{D}},z) - \ell(f_{\mathcal{D}^{\backslash i}},z)|\right] \leq \frac{2\rho\tau}{m} \mathbf{E}_{\mathcal{D},z}\left[\frac{1}{\lambda_1(\vec{H}(\hat{\theta}))}\right]\leq \frac{2\rho\tau}{m\inf\limits_{\|v\|=1}v^T\vec{H}(\hat{\theta})v},
    \label{eq:lr_hs}
\end{equation}
and for pointwise hypothesis stability we have
\begin{equation}
    \mathbf{E}_{\mathcal{D}}\left[|\ell(f_{\mathcal{D}},z_i) - \ell(f_{\mathcal{D}^{\backslash i}},z_i)|\right] \leq \frac{2\rho\tau}{m} \mathbf{E}_{\mathcal{D}}\left[\frac{1}{\lambda_1(\vec{H}(\hat{\theta}))}\right]\leq \frac{2\rho\tau}{m\inf\limits_{\|v\|=1}v^T\vec{H}(\hat{\theta})v}.
    \label{eq:lr_phs}
\end{equation}
Equations~\eqref{eq:lr_hs}~and~\eqref{eq:lr_phs} conclude the proof.
\qed
\end{myproof}
\label{thm:lr_hypothesis_stability}
\end{mytheorem}

Theorems~\ref{thm:lrl2_hypothesis_stability}~and~\ref{thm:lr_hypothesis_stability} are a generalization with respect to the loss function $\ell$. The cross-entropy loss $\ell(f_{\mathcal{D}},z)=y\log p(y|\vec{x};\theta) + (1-y)\log (1-p(y|\vec{x};\theta)$ and the $\ell_{\gamma}$ loss for $\gamma=1$ are both $1$-Lipschitzian, i.e., $\tau=1$.
On one hand, $L_2$ regularization stabilizes logistic regression since the penalty $\lambda$ is controllable. On the other, logistic regression is an unstable learning algorithm since $\inf_{\|v\|=1}v^T\vec{H}(\hat{\theta})v$ depends on $\mathcal{D}$ and the value of $\mathcal{L}_{\mathcal{D}}$ at $\hat{\theta}$. The eigenvector corresponding to the smallest possible eigenvalue of the Hessian $\vec{H}(\hat{\theta})$ of the cross-entropy loss at its minimizer $\hat{\theta}$ gives the direction of the smallest change, while the eigenvalue gives the magnitude of the change. This pair mostly depends on the input data in $\mathcal{D}$ and the eigenvalue can be very small and close to zero ($\inf_{\|v\|=1}v^T\vec{H}(\hat{\theta})v<<m$) when the gradient of the cross-entropy loss is very close to zero. This makes logistic regression highly unstable, hence the advantage of using $L_2$-regularization.

\section{Upper bounds on the generalization error of logistic regression and decision trees}
\label{sec:upper_bounds}

In this section, we give the upper bounds on the generalization error of decision trees, $L_2$-regularized logistic regression, and logistic regression. In that sense, we use the bounds from~\cite{elisseeff1}, based on hypothesis and pointwise hypothesis stability. Here, as previously, we consider a $\tau$-Lipschitzian loss $\ell.$

\begin{cor}[Hypothesis stability generalization upper bounds for decision trees]
Let $\beta_{h}(v)=1/v$ be the hypothesis stability of decision trees. Then the hypothesis stability upper bound on the generalization error of decision trees for any $\delta\in(0,1)$ and any $m \geq 1$ is
\begin{equation*}
    R_{gen}(f_{\mathcal{D}}) \leq R_{emp}(f_{\mathcal{D}}) + \frac{2}{v} + \left(\frac{4m}{v} + \tau\right)\sqrt{\frac{\log(1/\delta)}{2m}}
    \label{eq:upper_bounds_h_dt}
\end{equation*}
and holds with probability at least $1-\delta$ over the random draw of the training set $\mathcal{D}$.
\label{thm:upper_bounds_h_dt}
\end{cor}

\begin{cor}[Pointwise hypothesis stability generalization upper bounds for decision trees]
Let $\beta_{ph}(v)=1/v$ be the hypothesis stability of a decision tree with $v$ leaf nodes whose outcome is $f_{\mathcal{D}}$. Then the pointwise hypothesis stability upper bound on the generalization error of decision trees is
\begin{equation*}
    R_{gen}(f_{\mathcal{D}}) \leq R_{emp}(f_{\mathcal{D}}) + \frac{2}{v} + \left(\frac{4m}{v} + \tau\right)\sqrt{\frac{\log(1/\delta)}{2m}}
    \label{eq:upper_bounds_ph_dt}
\end{equation*}
and holds with probability at least $1-\delta$ over the random draw of the training set $\mathcal{D}$.
\label{thm:upper_bounds_ph_dt}
\end{cor}

\begin{cor}[Hypothesis and pointwise hypothesis stability generalization upper bounds for logistic regression]
Let $\beta_h(m)=2\rho\tau/m\inf_{\|v\|=1}v^T\vec{H}(\hat{\theta})v$ be the hypothesis (respectively pointwise hypothesis) stability of logistic regression whose outcome is $f_{\mathcal{D}}$. Then the hypothesis (respectively pointwise hypothesis) stability upper bound on the generalization error of logistic regression for any $\delta\in(0,1)$ and any $m\geq 1$ is
\begin{equation*}
    R_{gen}(f_{\mathcal{D}}) \leq R_{emp}(f_{\mathcal{D}}) + \frac{4\rho\tau}{m\inf\limits_{\|v\|=1}v^T\vec{H}(\hat{\theta})v} + \left(\frac{8\rho\tau}{\inf\limits_{\|v\|=1}v^T\vec{H}(\hat{\theta})v} + \tau\right)\sqrt{\frac{\log(1/\delta)}{2m}}
    \label{eq:upper_bounds_h_lr}
\end{equation*}
and holds with probability at least $1-\delta$ over the random draw of the training set $\mathcal{D}$.
\label{thm:upper_bounds_h_lr}
\end{cor}

\begin{cor}[Hypothesis, pointwise hypothesis, and uniform stability generalization upper bounds for $L_2$-regularized logistic regression]
Let $\beta_h(m)=2\rho\tau/\lambda m$ be the hypothesis (respectively pointwise hypothesis and uniform) stability of $L_2$-regularized logistic regression with penalty $\lambda > 0$ and outcome $f_{\mathcal{D}}$. Then the hypothesis (respectively pointwise hypothesis and uniform) stability upper bound on the generalization error of $L_2$-regularized logistic regression for any $\delta\in(0,1)$ and any $m \geq 1$ is
\begin{equation*}
    R_{gen}(f_{\mathcal{D}}) \leq R_{emp}(f_{\mathcal{D}}) + \frac{4\rho\tau}{\lambda m} + \left(\frac{8\rho\tau}{\lambda} + \tau\right)\sqrt{\frac{\log(1/\delta)}{2m}}
    \label{eq:upper_bounds_h_l2lr}
\end{equation*}
and holds with probability at least $1-\delta$ over the random draw of the training set $\mathcal{D}$.
\label{thm:upper_bounds_h_l2lr}
\end{cor}

\section{Stability measuring framework}
\label{sec:framework}

In this section, we briefly introduce a framework to measure the hypothesis and pointwise hypothesis stability of a learning algorithm. Our framework is based on the one introduced in~\cite[pages 6-14]{philippMeasuring}. With our framework, one can measure the hypothesis stability of a learning algorithm by estimating the expectation $\mathbf{E}_{\mathcal{D},z}\left[\ell(f_{\mathcal{D}}, z)-\ell(f_{\mathcal{D}^{\backslash i}}, z)\right]$ and pointwise hypothesis stability by estimating the expectation $\mathbf{E}_{\mathcal{D}}\left[\ell(f_{\mathcal{D}}, z_i)-\ell(f_{\mathcal{D}^{\backslash i}},z_i)\right]$. To that end, in the first part of this section, we describe the procedure to estimate the expected loss difference with respect to $\mathcal{D}$ and $z$, or only $\mathcal{D}$. In the second part, we define a procedure to measure the hypothesis stability, and, in the third, we define the procedure for estimating the pointwise hypothesis stability. In the last part of this section, we float the complete procedure.

\subsection{Estimating the expected loss difference}
In hypothesis stability, the absolute loss difference is averaged over all training sets $\mathcal{D}$ and all examples $z$ when the $i$-th example is removed from the training set. According to Definition~\ref{def:hypothesis_stability}, the expected loss difference has to be smaller than $\theta$ when any training example $z_i$ is removed. In other words, we need to take the largest expected absolute difference with respect to $i$.  In reality, however, we have only one training set available and we need to generate many replicates of $\mathcal{D}$ to estimate the expectation. For this, $B$ training sets $(\mathcal{D}_b)_{b=1}^B$ are resampled from $\mathcal{D}$ with bootstrap sampling (sampling with replacement). The simplest method of estimating the expectation is by averaging the absolute loss difference across the $B$ bootstrap samples, where $\Phi(\mathcal{D})$ is a function that depends on $\mathcal{D}$:

\begin{equation}
     \hat{\mathbf{E}}_{\mathcal{D}}\left[\Phi(\mathcal{D})\right] = \frac{1}{B} \sum_{b=1}^B \Phi(\mathcal{D}_b).
     \label{eq:est_exp_d}
\end{equation}

\subsection{Measuring hypothesis stability}
\label{sec:hsm}

As the introductory paragraph states, to measure the hypothesis stability of a learning algorithm with outcome $f_{\mathcal{D}}(\vec{x})$, we need to estimate the expected loss difference with respect to $\mathcal{D}$ and $z$.

We use Equation~\eqref{eq:est_exp_d} to estimate the expectation with respect to $\mathcal{D}$. Then, using the same approach from~\cite{philippMeasuring}, we generate an evaluation set $\mathcal{D}_{eval}$ that helps estimate the expectation with respect to all examples $z$. The evaluation set can be generated in three ways~\citep{philippMeasuring}:

\begin{itemize}
    \item[(1)] Out-of-bag (OOB): The evaluation set consists of all examples that belong to neither of the $B$ training sets ($\mathcal{D}_{eval}=\mathcal{D}\,\backslash\, \mathcal{D}_1 \cap \mathcal{D}_2 \cap \ldots \cap \mathcal{D}_B$).\\
    \item[(2)] Out-of-sample (OOS): The evaluation set is a completely new set of examples neither of which belongs to the original training set ($\mathcal{D}_{eval} \cap \mathcal{D} = \emptyset $). Since the all the available data for learning is in $\mathcal{D}$, we simulate this by splitting $\mathcal{D}$ into two sets $\mathcal{D}_{eval}$ and $\mathcal{D} \,\backslash\, \mathcal{D}_{eval}$.\\
    \item[(3)] All examples (ALL): the evaluation set comprises all $m$ training examples ($\mathcal{D}_{eval}=\mathcal{D}$).
\end{itemize}

\noindent
We estimate the expected loss difference with respect to $z$ by taking the mean over the evaluation set $\mathcal{D}_{eval}$:

\begin{equation}
    \hat{\mathbf{E}}_{z}[\Phi(z)] = \frac{1}{m'}\sum_{z' \in \mathcal{D}_{eval}}\Phi(z'),
    \label{eq:est_exp_z}
\end{equation}
where $m'=|\mathcal{D}_{eval}|$.

\noindent
By combining Equations~\eqref{eq:est_exp_d}~and~\eqref{eq:est_exp_z}, we get the following estimate of the hypothesis stability $\mathbf{E}_{\mathcal{D},z}\left[|\ell(f_{\mathcal{D}},z) - \ell(f_{\mathcal{D}^{\backslash i}},z)|\right]\leq\beta_h(m)$:

\begin{equation}
    \hat{\beta}_h(m) = \max_{i}\frac{1}{Bm'}\sum_{b=1}^B \sum_{z'\in\mathcal{D}_{eval}}|\ell(f_{\mathcal{D}_b},z') - \ell(f_{\mathcal{D}_b^{\backslash i}},z')|. 
    \label{eq:hypothesis_stability_measure}
\end{equation}

\subsection{Measuring pointwise hypothesis stability}

To measure pointwise hypothesis stability of a learning algorithm whose outcome is $f_{\mathcal{D}}(\vec{x})$, we need to estimate the expected loss difference over all training sets $\mathcal{D}$. Here, instead of an evaluation set $\mathcal{D}_{eval}$, we take the average loss difference at each example $z_i$ in the training set $\mathcal{D}$. Thus, $\mathcal{D}_{eval} = \mathcal{D}$, and to estimate the expectation $\mathbf{E}_{\mathcal{D}}\left[|\ell(f_{\mathcal{D}}, z_i) - \ell(f_{\mathcal{D}^{\backslash i}}, z_i)|\right] \leq \beta_{ph}(m)$ using Equation~\eqref{eq:est_exp_d}, we get the following measure for pointwise hypothesis stability:

\begin{equation}
    \hat{\beta}_{ph}(m) = \max_{i} \frac{1}{B}\sum_{b=1}^B |\ell(f_{\mathcal{D}},z_i) - \ell(f_{\mathcal{D}^{\backslash i}},z_i)|.
    \label{eq:pointwise_hypothesis_stability_measure}
\end{equation}

\subsection{Stability measuring procedure}
\label{sec:procedure}

In this part, we give a general procedure to measure the hypothesis and pointwise hypothesis stability of a learning algorithm. 

\begin{itemize}
    \item[(1)] Resample the training set $\mathcal{D}$ $B$ times with replacement to generate $B$ bootstrap samples $\mathcal{D}_1,\mathcal{D}_2,\ldots ,\mathcal{D}_B$.\\
    \item[(2)] Train one model on $\mathcal{D}$ and one model on each bootstrap sample $\mathcal{D}_1,\mathcal{D}_2,\ldots,\mathcal{D}_B$ using the same learning algorithm. This step yields $B +1$ trained models.\\
    \item[(3)] For hypothesis stability, generate an evaluation set $\mathcal{D}_{eval}$ using one of the methods from Section~\ref{sec:hsm} (OOB, OOS, or ALL). For pointwise hypothesis stability, take $\mathcal{D}_{eval}=\mathcal{D}$.\\
    \item[(4)] Compute $\hat{\beta}_h(m)$ using Equation~\eqref{eq:hypothesis_stability_measure} or $\hat{\beta}_{ph}(m)$ using Equation~\eqref{eq:pointwise_hypothesis_stability_measure}.
\end{itemize}

\section{Experiments}
\label{sec:experiments}

In this section, we confirm the results of this paper from Section~\ref{sec:base_stability}. To assess the theoretical results, we use the Hastie dataset for binary classification. The dataset has 10 Gaussian standard independent Gaussian features $X_1,\ldots,X_{10}$, and the target $Y$ is defined as
\begin{equation*}
    Y = \begin{cases}
                1 \quad \text{if } \sum_{j=1}^{10} X_j^2 > \chi_{10}^2(0.5),\\
                0\quad \text{otherwise},
                \end{cases}
\end{equation*}
where $\chi_{10}^2=9.34$ is the median of the chi-squared random variable with 10 degrees of freedom, which is equal to the sum of 10 standard Gaussians. The dataset, by default,  contains 2000 training examples and 10000 test examples, spread equally among the two classes. 
This dataset was borrowed from~\cite[p.339]{friedman2001elements}. 

To measure the hypothesis and pointwise hypothesis stability we employ the stability measuring procedure, given in Section~\ref{sec:procedure}. We repeat the four steps 10 times for $B=30$, that is, we use 30 bootstrap samples from the training set to estimate the expected absolute loss difference for both types of stability. 
To emphasize the effect of changes in the training set on the outcome of the learning algorithms, we reduced the dataset size to only 20 training examples ($m=20$). We constructed the evaluation set using the ALL strategy, that is, it consists of the same 20 training examples.
The reason for this is that removing or replacing an example from a large training set (e.g., 2000 examples) has little to no effect on the outcome of any learning algorithm. To summarize, we repeat steps (1) through (4), given in Section~\ref{sec:procedure}, 10 times, with $B=30$ and $m=20$ training examples in the Hastie dataset. Regarding step (3), we use the ALL strategy (see Section~\ref{sec:hsm}), i.e., we estimate the expected absolute loss difference using all training examples. 

For each learning algorithm, we measured hypothesis and pointwise hypothesis stability with respect to the variables on which they depend. We used the classification loss given in Section~\ref{sec:prerequisites}. We fixed the training set size at $m=20$ in all cases since it is already known that a larger $m$ always improves stability. To that end, we varied the number of leaves $v$ for decision trees, the penalty $\lambda$ for $L_2$-regularized logistic regression, and the number of gradient descent iterations to obtain different smallest eigenvalues of the Hessian matrix for logistic regression. To summarize, our experimental setup is as follows:

\begin{itemize}
    \item[$\bullet$] Size of the training set: $m=20$
    \item[$\bullet$] Data dimensionality: 10 Gaussians
    \item[$\bullet$] Bootstrap samples used to estimate the expected values: $B=30$
    \item[$\bullet$] Evaluation set generation: ALL, i.e., the whole training set
    \item[$\bullet$] Measures: $\hat{\beta}_h(20)$ and $\hat{\beta}_{ph}(20)$
    \item[$\bullet$] Number of trials: 10
\end{itemize}

\noindent
The values for $\hat{\beta}_h(20)$ and $\hat{\beta}_{ph}(20)$ are given in Table~\ref{tab:results} for decision trees, $L_2$-regularized logistic regression, and logistic regression, respectively. $\hat{\beta}_h(20)$ and $\hat{\beta}_{ph}(20)$ were measured 10 times and were then averaged.

\begin{table}[htb]
\begin{tabular*}{\textwidth}{l@{\extracolsep{\fill}}cclclcclc}
\hline
\multicolumn{3}{c}{\textbf{Decision tree}} & \multicolumn{3}{c}{\textbf{$L_2$-regularized LR}} & \multicolumn{3}{c}{\textbf{LR}} \\ \hline
$v$ & $\hat{\beta}_h(20)$ & $\hat{\beta}_{ph}(20)$ & $\lambda$ & $\hat{\beta}_h(20)$ & $\hat{\beta}_{ph}(20)$ & $T$ & $\hat{\beta}_h(20)$ & $\hat{\beta}_{ph}(20)$ \\ \hline
8 & 0.0483 & 0.9000 & 0.01 & 0.4183 & 0.1405 & 2 & 1.000 & 0.3333 \\
16 & 0.0431 & 0.7567 & 1.0 & 0.3216 & 0.0293 & 10 & 0.7867 & 0.2434 \\
64 & 0.0421 & 0.6300 & 2.0 & 0.2525 & 0.0202 & 50 & 0.6833 & 0.2172 \\
128 & 0.0373 & 0.6133 & 5.0 & 0.2235 & 0.0174 & 200 & 0.3233 & 0.1428  \\
256 & 0.0255 & 0.5500 & 10.0 & 0 & 0.0141 & 500 & 0.2767 & 0.0168 \\ \hline
\end{tabular*}
 \label{tab:results}
\caption{Hypothesis and pointwise hypothesis stability measured for decision trees, $L_2$-regularized and non-regularized logistic regression (LR). $T$ is the number of gradient descent iterations. Stability was measured for the classification loss. The values were average over ten trials.}
\end{table}

The results show that when the variable defining the hypothesis or pointwise hypothesis stability is changed so that it yields a smaller upper bound on the expected loss difference. Therefore, for decision trees, when the number of leaves $v$ is large, the expected loss difference decreases. Similarly, for $L_2$-regularized logistic regression, a larger penalty $\lambda$ improves stability. The same pattern holds for logistic regression as the smallest eigenvalue of the Hessian matrix decreases as the loss function is minimized.

\section{Conclusion and future work}
\label{sec:conclusion}
In this paper, we derived hypothesis and pointwise hypothesis stability of decision trees, logistic regression, and $L_2$-regularized logistic regression. The results presented here are the first look into how hypothesis and pointwise hypothesis stability interact in decision trees and logistic regression.
We also showed that logistic regression is not uniformly stable, while $L_2$-regularized logistic regression is. We showed that in all three cases, the stability depends on the size $m$ of the training set, and for logistic regression, it additionally depends on the penalty, i.e., the smallest eigenvalue of the Hessian matrix of the cross-entropy loss. Logistic regression cannot be considered a stable learning algorithm since its stability depends on an uncontrollable parameter. Using these results, we provided upper bounds on the generalization error of all three learning algorithms.

Moreover, we introduced a framework that allows one to precisely measure hypothesis and pointwise hypothesis stability using a training set. We then applied this framework to decision trees, logistic regression, and $L_2$-regularized logistic regression to measure the two kinds of stability using a synthetic 10-dimensional Gaussian dataset. The experiments showed that both kinds of stability increase along with the parameters that define them.

In the future, we anticipate that it is possible to craft new methods for training a logistic regression model that will additionally take the stability expressions into account. For instance, one can either include an additional term in the cost function of logistic regression or constrain the optimization problem in order to maximize the smallest possible eigenvalue of the Hessian matrix of the cross-entropy loss, $\inf_{\|v\|=1}v^T\vec{H}(\hat{\theta})v$. It would be challenging to estimate the Hessian in each step of the optimization process, and thus, we plan to employ different approximation techniques to that end.

\bibliographystyle{unsrt}

\end{document}